\documentclass[a4paper,10pt,twoside]{article}

\usepackage{clin}        %
\usepackage{harvard}     %

\usepackage{subcaption}
\usepackage{booktabs}
\usepackage{hyperref}
\usepackage{todonotes}
\usepackage{layouts}

\begin{document}
\title{Measuring Shifts in Attitudes Towards COVID-19 Measures in Belgium Using Multilingual BERT}

\author{Kristen Scott$^*$ \email{kristen.scott@cs.kuleuven.be}\\
{\normalsize \bf Pieter Delobelle}$^*$ \email{pieter.delobelle@cs.kuleuven.be}\\
{\normalsize \bf Bettina Berendt}$^{**}$ \email{bettina.berendt@cs.kuleuven.be}
\AND \addr{$^*$Department of Computer Science; Leuven.AI, KU Leuven, Belgium\\Also denotes equal contribution}
\AND \addr{$^{**}$TU Berlin and Weizenbaum Institute, Germany; KU Leuven, Belgium}}

\maketitle\thispagestyle{empty} %

\begin{abstract}
We classify seven months' worth of Belgian COVID-related Tweets using multilingual BERT and relate them to their governments' COVID measures.
We classify Tweets by their stated opinion on Belgian government curfew measures (too strict, ok, too loose). We examine the change in topics discussed and views expressed over time and in reference to dates of related events such as implementation of new measures or COVID-19 related announcements in the media. 
\end{abstract}

\section{Introduction and Related Work}

Sentiment analysis or opinion mining of social media content presents the possibility of following trends in public discussion. Twitter, with an easy to use API and short, focused messages called Tweets, is often targeted for such tasks \cite{medhat2014sentiment-analysis,giachanou2016twitter-sentiment-analysis}. During the COVD-19 pandemic, quantifying which measures are supported by the general population, and which ones are not, could be useful in shaping the course of a nation's strategy. 
Recent work has focused on monitoring reactions to the COVID pandemic utilizing sentiment analysis  \cite{wang_covid-19_2020,chen2020corona-dataset,brandl2020corona-dataset,kurten2021belgian-corona-dataset,wang_public_2020}. However, sentiment does not necessarily map to opinions on more complex opinions about measures. \citeasnoun{wang_public_2020} presented initial results in workshop on classifying stances (for or against) towards Dutch government policies on masks and distancing using a neural network. Others have incorporated qualitative analysis techniques into similar work flows in an attempt to gain a more nuanced understanding of social media discussion than sentiment analysis, unsupervised machine learning or classification models alone \cite{jimenezsotomayor_coronavirus_2020,xue_public_2020}.

BERT~\cite{devlin2019} is a Natural Language Processing (NLP) model that uses pre-training and fine-tuning. 
This makes it easier than ever to create custom, domain-adapted classifiers that capture the nuances of discussions and opinions expressed in a given domain. 
We characterize the discussion of Belgian COVID measures on Twitter over time using multilingual BERT models that we finetuned on manually labelled Tweets. 
By visualizing the change of rate of the sentiment in curfew-related tweets, we found shifts in the support for curfews over time.

\section{Methodology}
We used a multilingual BERT model to classify 1.3 million Tweets related to the COVID-19 pandemic, based on a manually labeled training set, as described in \autoref{ss:labeling}.
The Tweets were collected through the Twitter API starting from October 13, 2020 until April 08, 2021 using a continuously running script\footnote{This script with search terms and all our code is available at \url{https://github.com/iPieter/bert-corona-tweets}.}. 
Tweets were collected using (i) multilingual search terms related to COVID-19, corona and specific related topics, (ii) a language filter on Dutch, French and English, and (iii) a filter for locations in Belgium.

In \autoref{ss:training}, we describe how we use this dataset and the collected labels to develop multiple models. 
These models were developed synchronously with the labeling task.
We started with model to filter irrelevant Tweets (e.g. news announcements) to save labeling time (Sieve I).
We then created a second model to classify the Tweets into topics, which we use to focus on the curfew topic (Sieve II)
for the more challenging labels: measure and government support.
This interplay allowed us to reduce labeling cost and develop multiple useful models.

\begin{table}[t]
\small
  \centering
  \caption{Labeling categories for each tweet.}\label{tab:axes}
  \resizebox{0.86\textwidth}{!}{%

 \begin{tabular}{ccccc} 
 \toprule
 \multicolumn{2}{c}{\textsc{Topic}} & \textsc{Measure Support} & \textsc{Government Support} &  \textsc{Relevance}\\
 \midrule
 masks      &testing  &   too-strict & supportive & irrelevant\\ 
 curfew     &closing-horeca &  ok & unsupportive  & \\
 quarantine &vaccine  &   too-loose  & not-applicable  & \\
 lockdown   &other-measure & not-applicable  &  & \\
 schools    &  &   &  \\

 \bottomrule
\end{tabular}
}
\end{table}
\subsection{Labeling}\label{ss:labeling}
Two manually labeled datasets were used for training. The first, consisting of 1695 Tweets was used for classifying topics. The second set of 2000 labeled Tweets was used to classify support for curfews. As described in \autoref{tab:axes}, Tweets were labeled by topic (curfew measure), as well as by two opinion axis: opinion toward specific measures (too strict, acceptable, not strict enough and a neutral option) and measure of the overall support expressed towards the government's handling of the pandemic (supportive, unsupportive).

We developed a code book which defines the labels  procedure in detail. This labeling process was tested and refined through two rounds of labeling on smaller datasets with Belgian and multilingual labelers. Each round was followed by discussion, resolving of disagreement and making minor adjustments to the code book \footnote{Code book available at \url{https://github.com/iPieter/bert-corona-tweets}.}.

\subsection{Training}\label{ss:training}
\begin{figure}
    \centering
    \caption{Schematic illustration of our contribution with domain-specific classifiers.}
    \label{fig:pipeline}
    \includegraphics[width=0.88\textwidth]{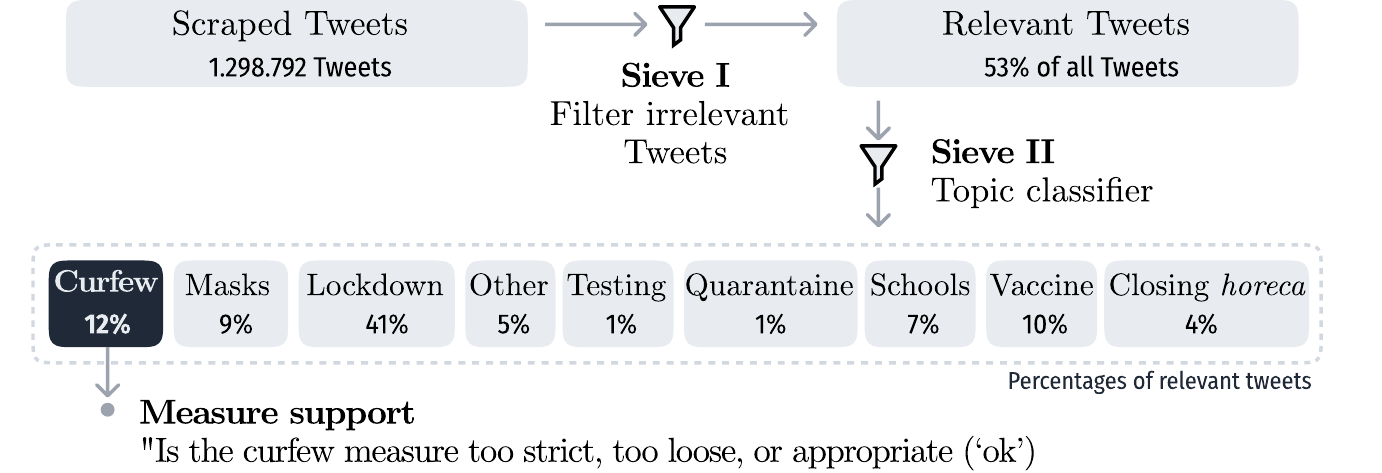}

\end{figure}

We developed multiple models to classify Tweets, which correspond with the labeling rounds.
\autoref{fig:pipeline} shows how collected Tweets on the topics are filtered and that we have four models: (i) a model to filter irrelevant Tweets for Sieve I, (ii) another model to classify topics and two models to (iii) predict support for a measure and (iv) support for the government. 
As mentioned before, each sieve helped reduce the number of Tweets that needed to be classified each round.

\paragraph{Classifying relevant Tweets.}
For the first sieve, we focus on relevant versus irrelevant Tweets as discussed in \autoref{ss:labeling}
Only 53\% of the labeled Tweets were relevant.
To automatically filter these Tweets, we trained and evaluated multilingual BERT (mBERT) and XLM-RoBERTa models. 
Each training was run 8 times with random hyperparameters and the best-performing model---using accuracy as a selection metric---was evaluated on a held-out test set, following \citeasnoun{dodge-etal-2019-show}.

mBERT performed slightly better than XLM-RoBERTa, with an AUC score of 0.85 and 0.84 respectively. 
The mBERT model also had a higher true positive rate of 0.3 when selecting a threshold with a false positive rate of zero, as the goal of the first sieve is to remove \emph{clearly} irrelevant Tweets.
From a computational standpoint, the base mBERT model also has the benefit that it is significantly cheaper and faster to train due to a smaller model size.

\paragraph{Classifying topics.}
We trained mBERT on 600 labeled Tweets to classify topics, we validated 8 models on a validation set with 64 Tweets and finally tested the best-performing model, using accuracy, on 100 Tweets.
The best-performing model had an overall accuracy of 0.73 (macro-averaged accuracy of 0.95). 
Some classes perform very well, like {\tt curfew} ($AUC=0.90$), {\tt lockdown} ($AUC=0.85$) and {\tt vaccine} ($AUC=0.90$).
Yet, some classes are ill-represented in the dataset and perform significantly worse, more specifically {\tt quarantine} ($AUC=0.50$) and {\tt testing} ($N=1$).

The topic model performs quite well overall and given our interest in the curfew topic specifically, this model is quite suitable. 
We also make the model available on the HuggingFace repository\footnote{Available at \url{https://huggingface.co/DTAI-KULeuven/mbert-corona-tweets-belgium-topics}.} for practitioners to use.

\paragraph{Classifying support for curfews.}
For the last classification model, we trained mBERT for multiclass classification on 1518\footnote{This were originally 2000 Tweets of which the clearly irrelevant ones were filtered with Sieve I before labeling.} Tweets with support labels, of which 100 were used as held-out test set and 75 as validation split.
We tested 5 hyperparameter assignments.
The overall accuracy is 0.71. However, there is a significant class imbalance and despite oversampling, the performance varies from no better than random ($AUC=0.5$ for {\tt too-loose}) to good ($AUC=0.74$ for {\tt not-applicable}, $AUC=0.69$ for {\tt ok} and $AUC=0.73$ for {\tt too-strict}). 

Given these results, we primarily focus on the {\tt too-strict} label for the curfew topic in the rest of this work. 
We also make this model available through the HuggingFace repository\footnote{Available at \url{https://huggingface.co/DTAI-KULeuven/mbert-corona-tweets-belgium-curfew-support}.}.

\section{Results and discussion}

\begin{figure}[tb]
\centerline{\includegraphics[width=0.95\columnwidth]{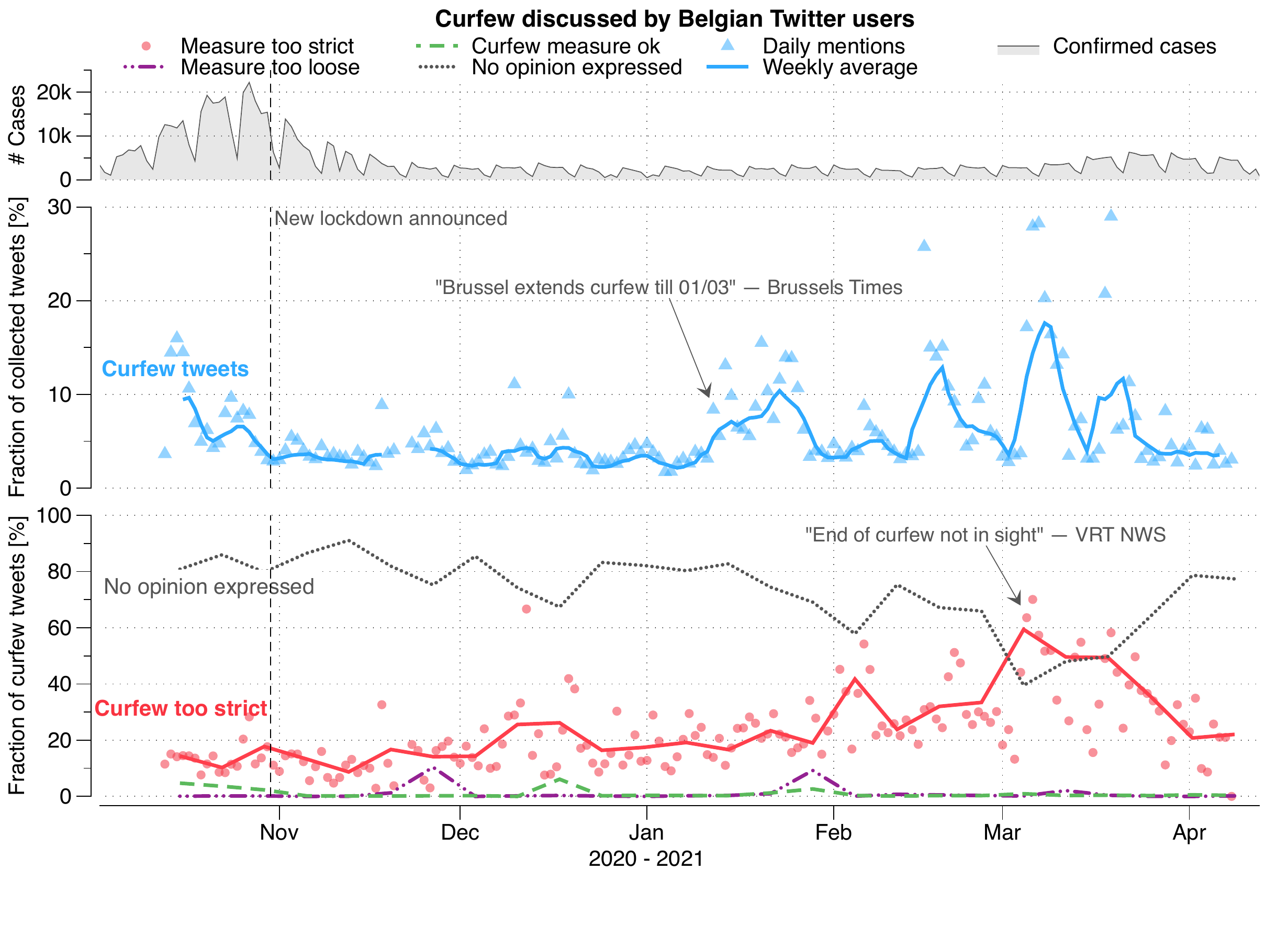}}
\caption{Timeline of the relative number of Tweets on the curfew topic (middle) and the fraction of those Tweets that find the curfew too strict, too loose, or a suitable measure (bottom), with the number of daily cases in Belgium to give context on the pandemic situation (top). }
\label{fig_timeline_curfew}
\end{figure}

Here we focus on the topic of curfew for reporting of more detailed results. The timeline of the rate of classified Tweets with the topic of curfew, along with classified rate of support (or non-support) for curfews, is shown in \autoref{fig_timeline_curfew}. Also included, for reference, is the rate of confirmed COVID cases in Belgium~\cite{sciensano_2021}.

November 2, 2020, Belgium entered a country-wide lockdown which included a national midnight curfew, while some regional curfews had been put in place in the days prior. We find media announcements of these upcoming curfews as well as announcements regarding the extension of these curfews \cite{nws_liveblog_nodate,johnston_brussels_2021} were accompanied by temporary increases curfew related Tweets. In October, as the rate of curfew Tweets dropped, there was no change in the opinions expressed about the curfew (with the majority remaining 'no opinion' until February). By contrast, during the 2021 increases in curfew Tweets we see a large change in opinions (particularly an increase in `too strict'). Further research is required to determine whether the changes in rate of negative opinion observed correspond to changes in public opinion or some other effect such as increased attention to particular announcements by individuals with consistent anti-curfew opinions. 

We also see that opinions on measure strictness are just one element of the discussions around COVID measures, suggesting the use of Twitter for other forms of communication, such as information sharing and humour as well as conveying complex points of views and personal stories (e.g. about the impact of the curfew). The ability of the BERT models to classify tweets based on our highly specific scale of strictness suggests that such models may be effective for categorizing based on other complex and nuanced labels when trained with carefully labeled data.

\section{Conclusion and future work}

We are able to observe the discussion of COVID measures on Twitter over time examine the shifting in both the topics of focus and the view points. We found that the majority of Twitter discussion of these measures is not centered on expression of specific levels of support and thus identify the need to characterize the nature of the non-opinionated Tweets as well as to understand the poorer performance of our model on opinions other than `no opinion' and `too strict'. We also acknowledge the limitations of treating Twitter data as representative of the viewpoints of the general population. While we do work with multiple languages, further work can be done to determine differential performance between languages, dialects and informal and slang texts.

\section*{Acknowledgment}
Kristen Scott was supported by the NoBIAS — H2020-MSCA-ITN-2019 project GA No. 860630. Pieter Delobelle was supported by the Research Foundation - Flanders (FWO) under EOS No. 30992574 (VeriLearn). Pieter Delobelle also received funding from the Flemish Government under the ``Onderzoeksprogramma Artificiële Intelligentie (AI) Vlaanderen'' programme.

\bibliographystyle{clin} 
\bibliography{paper}  

\end{document}